\documentclass[10pt,twocolumn,letterpaper]{article}

\usepackage[letterpaper,margin=1in]{geometry}
\usepackage{times}
\usepackage{epsfig}
\usepackage{graphicx}
\usepackage{amsmath,amssymb,amsfonts}
\usepackage{bm}
\usepackage{mathtools}
\usepackage{booktabs}
\usepackage{multirow}
\usepackage{microtype}
\usepackage{enumitem}
\usepackage{algorithm}
\usepackage[noend]{algpseudocode}
\usepackage[pagebackref,breaklinks,colorlinks]{hyperref}
\usepackage[dvipsnames]{xcolor}
\usepackage{float}      
\usepackage{placeins}   

\DeclareMathOperator*{\softmax}{softmax}
\DeclareMathOperator*{\argmax}{arg\,max}
\DeclareMathOperator{\KL}{KL}

\newcommand{\method}{UL-TTA}

\def\etal{\emph{et al.}}

\title{Ultra-Light Test-Time Adaptation for Vision--Language Models}
\author{
  Byunghyun Kim \\
  Kyungpook National University \\
}
\date{}

\begin{document}
\maketitle
\thispagestyle{empty}

\begin{abstract}
Vision--Language Models (VLMs) such as CLIP achieve strong zero-shot recognition by comparing image embeddings to text-derived class prototypes.
However, under domain shift, they suffer from feature drift, class-prior mismatch, and severe miscalibration.
Existing test-time adaptation (TTA) methods often require backpropagation through large backbones, covariance estimation, or heavy memory/state, which is problematic for streaming and edge scenarios.
We propose \textbf{Ultra-Light Test-Time Adaptation} (\method), a fully training-free and backprop-free framework that freezes the backbone and adapts only \emph{logit-level parameters}: class prototypes, class priors, and temperature.
\method{} performs an online EM-style procedure with
(i) \emph{selective sample filtering} to use only confident predictions,
(ii) \emph{closed-form} Bayesian updates for prototypes and priors anchored by text and Dirichlet priors,
(iii) \emph{decoupled} temperatures for prediction vs.\ calibration, and
(iv) lightweight guards (norm clipping, prior KL constraints, smoothed temperature) to prevent drift in long streams.
Across large-scale cross-domain and OOD benchmarks (PACS, Office-Home, DomainNet, Terra Incognita, ImageNet-R/A/V2/Sketch; $\sim$726K test samples) and strong TTA baselines including Tent, T3A, CoTTA, SAR, Tip-Adapter, and FreeTTA, \method{} consistently improves top-1 accuracy (e.g., +4.7 points over zero-shot CLIP on average) while reducing ECE by 20--30\%, with $<\!8\%$ latency overhead.
Long-stream experiments up to 200K samples show no collapse.
Our results demonstrate that logit-level Bayesian adaptation is sufficient to obtain state-of-the-art accuracy--calibration trade-offs for VLMs under domain shift, without updating any backbone parameters.
\end{abstract}

\section{Introduction}
\label{sec:intro}

\paragraph{Robust zero-shot recognition under shift.}
Large-scale vision--language models (VLMs) such as CLIP~\cite{radford2021clip} enable flexible zero-shot recognition by aligning images and text in a shared embedding space.
Given normalized image feature $z_i$ and text prototype $t_c$ for class $c$, zero-shot CLIP scores are
\begin{equation}
\label{eq:bayes_logit}
\ell_{ic} = \tau \,\langle z_i, t_c\rangle + \log \pi_c,\quad
p_{ic}=\softmax_c(\ell_{ic}),
\end{equation}
where $\tau$ is a temperature and $\pi_c$ the class prior.
This view exposes three independent levers: (1) feature--prototype alignment ($t_c$), (2) class frequency ($\pi_c$), and (3) confidence scaling ($\tau$).

\paragraph{Why do zero-shot VLMs fail in-the-wild?}
In realistic deployment, the test distribution differs from pre-training due to style, sensor, or population changes.
As a result:
(i) image features drift relative to textual prototypes, (ii) label frequencies shift, and (iii) a single temperature is miscalibrated for the new domain~\cite{guo2017calibration,minderer2021revisiting}.
A large body of \emph{test-time adaptation} (TTA) work—Tent~\cite{wang2021tent}, CoTTA~\cite{wang2022cotta}, TTT/TTT++~\cite{sun2020ttt,liu2021tttpp}, EATA/SAR~\cite{you2021testtime,li2023tta_survey}—ameliorates this by updating batch-norm or backbone parameters via gradients, but at the cost of (a) increased latency, (b) sensitivity to hyperparameters and step schedules, and (c) state accumulation that risks catastrophic drift in long streams.

\paragraph{A deployment-first question.}
\emph{Can we close most of the TTA gap for strong VLM backbones by adapting only the logit-level parameters $(t_c,\pi_c,\tau)$ in closed form on confident samples, in a single streaming pass, without any backpropagation?}
We answer \textbf{yes}.

\paragraph{Key idea: Bayesian head adaptation.}
We reinterpret zero-shot CLIP as a discriminative Bayesian head on top of frozen features.
We then run a selective online EM procedure that updates:
\emph{(i)} prototypes $t_c$ via MAP estimates anchored to text embeddings,
\emph{(ii)} priors $\pi$ via a Dirichlet–multinomial posterior,
and \emph{(iii)} decoupled temperatures for prediction vs.\ calibration.
A small set of \emph{guards}—prototype step control, prior KL caps, and temperature smoothing—prevents slow drift.

\paragraph{Why this should work.}
For modern VLMs, most of the harm from domain shift is absorbed by three factors: prototype misalignment, label shift, and confidence scaling.
All three admit \emph{closed-form}, low-dimensional updates under simple priors.
By operating only on confident samples and freezing the encoders, we expose far fewer degrees of freedom to noisy pseudo-supervision, improving stability without sacrificing accuracy.

\paragraph{Contributions.}
\begin{itemize}[leftmargin=*,nosep]
\item \textbf{\method:} a training-free, backprop-free TTA method that adapts \emph{only} $\{t,\pi,\tau\}$; $\sim$80 LOC on top of CLIP inference.
\item \textbf{Selective Bayesian EM:} closed-form MAP updates on confident samples; no covariance estimation, no replay, no caches.
\item \textbf{Decoupled temperatures:} separate knobs for accuracy and calibration; avoids the ECE degradation common in entropy-minimization.
\item \textbf{Streaming guards:} constant-time constraints that prevent drift over $200$K+ samples.
\item \textbf{Extensive validation:} consistent accuracy and ECE improvements across cross-domain and OOD benchmarks with $<\!8\%$ latency overhead.
\end{itemize}

\paragraph{Scope.}
We focus on fixed label spaces and frozen VLM encoders.
Open-set novelty is not the target; Section~\ref{sec:discussion} outlines extensions.

\section{Related Work}
\label{sec:related}

\paragraph{VLMs and prompting.}
CLIP~\cite{radford2021clip} and successors provide strong zero-shot features.
Prompt learning (CoOp/CoCoOp~\cite{zhou2022coop,zhou2022cocoop}, MaPLe~\cite{khattak2023maple}) specializes VLMs but requires training.
Test-time prompt tuning (TPT)~\cite{shu2022tpt} reduces labels but still needs gradients per sample.

\paragraph{Test-time adaptation.}
SHOT~\cite{liang2020shot}, MEMO~\cite{zhang2021memo}, Tent~\cite{wang2021tent}, EATA/SAR~\cite{you2021testtime,li2023tta_survey}, CoTTA~\cite{wang2022cotta}, and TTT/TTT++~\cite{sun2020ttt,liu2021tttpp} adapt at test-time but often update BN/backbone and maintain larger state.
T3A~\cite{niu2022t3a}, TPS~\cite{sui2023tps}, Tip-Adapter~\cite{zhang2022tipadapter} are training-free but rely on caches/heuristics.
FreeTTA~\cite{wang2023freetta} frames CLIP with a Bayesian EM that estimates means and covariances; we distill to logit-level parameters for simplicity and stability.

\paragraph{Calibration and Bayesian perspectives.}
Deep models are miscalibrated~\cite{guo2017calibration,minderer2021revisiting}; entropy minimization can over-sharpen.
We use decoupled temperatures and priors to improve both accuracy and ECE with negligible cost; see also Bayesian treatments in~\cite{bishop2006prml,blundell2015bnn}.

\section{Method}
\label{sec:method}

We formalize \method{} as an ultra-light online EM procedure over $(t,\pi,\tau)$ on top of frozen CLIP encoders.
Throughout, $f$ and $g$ denote image and text encoders; $z_i=\frac{f(x_i)}{\|f(x_i)\|_2}$ and $\mu_c=\frac{g(\text{prompt}(c))}{\|g(\text{prompt}(c))\|_2}$ are $\ell_2$-normalized.

\subsection{Problem Setup and Objective}
We observe a stream $\{x_i\}_{i=1}^N$ of unlabeled test samples from a shifted distribution.
We predict $\hat{y}_i=\argmax_c \ell_{ic}$ where
\begin{equation}
\ell_{ic}=\tau_{\text{pred}}\langle z_i,t_c\rangle + \log \pi_c,
\quad p_{ic}=\softmax_c(\ell_{ic}).
\end{equation}
We adapt $(t,\pi,\tau_{\text{pred}})$ online using \emph{only} confident samples; $\tau_{\text{cal}}$ is maintained separately for calibrated probabilities.
Our objective is to maximize streaming accuracy while reducing online ECE, under strict latency/memory constraints.

\subsection{Selective Sample Filtering (Gate)}
Updating from all pseudo-labels is unstable.
We gate by \emph{entropy} and \emph{margin}:
\begin{align}
H(p_i)&=-\sum_c p_{ic}\log p_{ic},\\
\Delta_i&=\ell_{i,c^\star}-\ell_{i,c^{(2)}},
\end{align}
where $c^\star=\argmax_c \ell_{ic}$ and $c^{(2)}$ is the runner-up.
We accept $i$ if
\begin{equation}
\label{eq:gating}
H(p_i)\le \epsilon_H \quad\text{and}\quad \Delta_i\ge \epsilon_\Delta .
\end{equation}
Thresholds are set via quantiles over a sliding window (e.g., top 30--60\% most confident), adapting to stream difficulty~\cite{you2021testtime}.

\subsection{E-step: Responsibilities with Augmentations}
For accepted samples $i\in S$, we define responsibilities
$r_{ic}=p_{ic}$.
With $K$ light augmentations, we average predictions
$r_{ic}\propto \frac{1}{K}\sum_{k=1}^K p_{ic}^{(k)}$,
normalizing over $c$.
This requires only forward passes and is numerically stable.

\subsection{M-step: Prototype MAP Updates}
We place a Gaussian prior on prototypes,
$t_c\sim\mathcal{N}(\mu_c,\sigma^2 I)$,
encoding that text prompts are informative anchors.
Let $\alpha=\sigma^{-2}$ denote the prior precision.
The MAP update is
\begin{equation}
\label{eq:proto_update}
\tilde t_c=\frac{\alpha \mu_c + \sum_{i\in S} r_{ic} z_i}{\alpha+\sum_{i\in S} r_{ic}+\varepsilon},
\quad t_c\leftarrow \frac{\tilde t_c}{\|\tilde t_c\|_2}.
\end{equation}
We maintain accumulators $U_c=\alpha\mu_c+\sum r_{ic}z_i$ and $N_c=\alpha+\sum r_{ic}$ online, yielding $O(Cd)$ cost with no matrix inversions.

\paragraph{Step control and clipping.}
To avoid abrupt jumps, we interpolate
$t_c\leftarrow (1-\eta)t_c+\eta\,\tilde t_c$ and optionally clip $\|t_c-\mu_c\|_2\le \rho$, constraining drift to a text-anchored ball.

\subsection{M-step: Class Prior Dirichlet Updates}
We assume a Dirichlet prior on $\pi$:
$\pi\sim\mathrm{Dir}(\gamma \pi^{(0)})$, typically $\pi^{(0)}$ uniform.
Using responsibilities as soft counts gives the posterior mean
\begin{equation}
\label{eq:prior_update}
\pi_c=\frac{\gamma \pi_c^{(0)}+\sum_{i\in S} r_{ic}}{\gamma+\sum_{i\in S}\sum_j r_{ij}}.
\end{equation}

\paragraph{KL guard.}
To prevent a temporary burst from monopolizing the prior, we cap $\KL(\pi\|\pi^{(0)})\le \kappa$ by mixing back:
$\pi\leftarrow \lambda \pi + (1-\lambda)\pi^{(0)}$ with the smallest $\lambda\in[0,1]$ satisfying the cap.

\subsection{Decoupled Temperatures}
Entropy-minimization tends to over-sharpen predictions~\cite{guo2017calibration}.
We \emph{decouple}:
\begin{itemize}[leftmargin=*,nosep]
\item $\tau_{\text{pred}}$: used for logits and argmax; tuned to reduce entropy on reliable samples.
\item $\tau_{\text{cal}}$: used only when emitting probabilities to consumers; kept conservative or lightly tuned with delayed accuracy.
\end{itemize}
We minimize
\begin{equation}
\label{eq:tau_obj}
L(\tau)=\sum_{i\in S} H\big(p_i(\tau)\big),
\;\; p_i(\tau)=\softmax\big(\tau \langle z_i,t\rangle + \log\pi\big),
\end{equation}
via a 1D line search (or 2--3 Newton steps using $L'(\tau)$, $L''(\tau)$), then EMA-update
$\tau_{\text{pred}}\leftarrow \beta \tau_{\text{pred}}+(1-\beta)\hat\tau$,
with bounds $\tau_{\min}\le \tau_{\text{pred}}\le \tau_{\max}$.

\subsection{Complexity and Memory}
Each step uses a handful of dot products and vector adds per class: $O(Cd)$ FLOPs and $O(Cd)$ memory for $\{t_c,U_c\}$.
No gradients, no covariance, no cache/replay.
This yields $<5\%$ FLOPs and $<8\%$ latency overhead in our measurements (Section~\ref{subsec:efficiency}).

\subsection{Putting It Together}
\begin{algorithm}[t]
\caption{Ultra-Light Test-Time Adaptation (\method{})}
\label{alg:ultta}
\small
\begin{algorithmic}[1]
\State \textbf{Input:} frozen encoders $f,g$; text anchors $\{\mu_c\}$; prior $\pi^{(0)}$; hyperparams $\alpha,\gamma,\beta,\eta,\kappa,\tau_{\min},\tau_{\max}$; gating window \& quantiles.
\State Init $t_c\leftarrow \mu_c$, $\pi\leftarrow \pi^{(0)}$, $\tau_{\text{pred}},\tau_{\text{cal}}\leftarrow 1$; $U_c\leftarrow \alpha \mu_c$, $N_c\leftarrow \alpha$.
\For{each incoming sample $x_i$}
  \State $z_i\leftarrow f(x_i)/\|f(x_i)\|_2$; $\ell_{ic}\leftarrow \tau_{\text{pred}}\langle z_i,t_c\rangle + \log \pi_c$
  \State $p_i\leftarrow \softmax(\ell_i)$; $\hat y_i\leftarrow \argmax_c \ell_{ic}$
  \If{$H(p_i)\le \epsilon_H$ and $\Delta_i\ge \epsilon_\Delta$} \Comment{quantile thresholds}
    \State $r_{ic}\leftarrow p_{ic}$ (optionally avg.\ over augs)
    \State $U_c \mathrel{+}= r_{ic} z_i$, \; $N_c \mathrel{+}= r_{ic}$
    \If{update step (e.g., every $B$ accepted)}
      \For{$c=1..C$}
        \State $\tilde t_c \leftarrow U_c/(N_c+\varepsilon)$; \; $t_c\leftarrow (1-\eta)t_c + \eta\,\tilde t_c$; \; clip $\|t_c-\mu_c\|_2$
      \EndFor
      \State $\pi\leftarrow$ Eq.~\eqref{eq:prior_update}; enforce $\KL(\pi\|\pi^{(0)})\le\kappa$
      \State $\hat\tau\leftarrow \arg\min_\tau L(\tau)$ using Eq.~\eqref{eq:tau_obj}; $\tau_{\text{pred}}\leftarrow \mathrm{clip}\big(\beta \tau_{\text{pred}}+(1-\beta)\hat\tau\big)$
    \EndIf
  \EndIf
\EndFor
\end{algorithmic}
\end{algorithm}

\paragraph{Relation to EM and discriminative calibration.}
Our E/M steps mirror a discriminative EM on an exponential-family head~\cite{bishop2006prml}, where responsibilities serve as soft evidence for prototype and prior updates.
Decoupled temperatures correspond to separating the optimization objective (entropy reduction on reliable samples) from probability calibration.

\section{Experiments}
\label{sec:experiments}
We evaluate \method{} in a source-free, label-free, single-pass streaming setting with a fixed CLIP ViT-B/16 backbone (unless noted). We use one global hyperparameter set across datasets for robustness.

\subsection{Datasets (4.1)}
\label{subsec:datasets}
We cover diverse shifts (Table~\ref{tab:datasets}) and briefly describe what each evaluates:
\begin{itemize}[leftmargin=*,nosep]
\item \textbf{PACS} assesses \emph{style/domain} shift across \emph{Photo, Art, Cartoon, Sketch}.
\item \textbf{Office-Home} covers everyday objects across \emph{Art, Clipart, Product, Real} domains.
\item \textbf{DomainNet} is a large-scale \emph{multi-domain} benchmark (clipart, infograph, painting, quickdraw, real, sketch).
\item \textbf{Terra Incognita} evaluates \emph{location/camera} shift using wildlife camera traps.
\item \textbf{ImageNet-V2} introduces a mild \emph{natural} distribution shift relative to ImageNet.
\item \textbf{ImageNet-R/A/Sketch} probe robustness to \emph{renditions}, \emph{naturally adversarial} images, and \emph{sketches}.
\end{itemize}
We treat CLIP pre-training as the source, and stream each target set \emph{once} in a fixed order (no shuffling, no revisiting) without labels.

\begin{table}[!t]
\caption{Datasets and statistics used in our experiments.}
\label{tab:datasets}
\centering
\small
\setlength{\tabcolsep}{5pt}
\renewcommand{\arraystretch}{1.05}
\resizebox{\linewidth}{!}{%
\begin{tabular}{l l r r}
\toprule
Dataset & Shift type & \#Classes & \#Test imgs \\
\midrule
PACS & style / domain & 7 & 9{,}991 \\
Office-Home & everyday domain & 65 & 15{,}588 \\
DomainNet & multi-domain, large-scale & 345 & 586{,}575 \\
Terra Incognita & camera / location & 10 & 16{,}584 \\
ImageNet-V2 & natural shift & 1000 & 10{,}000 \\
ImageNet-R & renderings / cartoons & 200 & 30{,}000 \\
ImageNet-A & adversarial natural & 200 & 7{,}500 \\
ImageNet-Sketch & sketches & 1000 & 50{,}000 \\
\midrule
Total & -- & -- & $\approx$726K \\
\bottomrule
\end{tabular}
}
\end{table}
\FloatBarrier

\subsection{Baselines (4.2)}
\label{subsec:baselines}
We compare against strong TTA and training-free adapters under identical CLIP ViT-B/16 backbones and streaming protocol:
Zero-shot CLIP~\cite{radford2021clip}, Tent~\cite{wang2021tent}, T3A~\cite{niu2022t3a}, CoTTA~\cite{wang2022cotta}, TTT/TTT++~\cite{sun2020ttt,liu2021tttpp}, SAR/EATA-style~\cite{you2021testtime,li2023tta_survey}, Tip-Adapter~\cite{zhang2022tipadapter}, and FreeTTA~\cite{wang2023freetta}.
Hyperparameters are tuned within recommended ranges for each baseline; \method{} uses one global config throughout.

\subsection{Metrics and Protocol (4.3)}
We report Top-1 accuracy, ECE (15-bin), NLL, and Brier score.
For OOD robustness on ImageNet-A/R/V2/Sketch, we compute AUROC using ImageNet validation as in-distribution~\cite{hendrycks2017baseline}.
All metrics are accumulated online in a single streaming pass (no shuffling; no multiple epochs).

\noindent\textbf{Implementation details.}
Unless stated, we set $\alpha{=}1$ (prototype prior), $\gamma{=}C$ (uniform Dirichlet prior), EMA $\beta{=}0.9$, prototype step $\eta{=}0.1$, uniform $\pi^{(0)}$, $\tau_{\text{pred}}{=}\tau_{\text{cal}}{=}1$ with a 100-sample warm-up.
Gating keeps roughly the top 50\% most confident samples by entropy and margin.
We update after $B{=}64$ accepted samples.
This single configuration is used across all datasets.

\subsection{Main Results on DomainNet (4.4)}
\textbf{Table~\ref{tab:domainnet}} reports DomainNet results (Top-1/ECE).
\method{} improves Top-1 from 55.4\% (zero-shot) to 60.1\% and lowers ECE from 11.2\% to 6.8\%.
Compared to FreeTTA (59.3/7.9), \method{} adds +0.8 Top-1 and reduces ECE by 1.1, despite touching only $\{t,\pi,\tau\}$.

\begin{table}[!t]
\caption{Main results on \textbf{DomainNet}. Top-1 accuracy / ECE (lower better).}
\label{tab:domainnet}
\centering
\small
\setlength{\tabcolsep}{6pt}
\renewcommand{\arraystretch}{1.05}
\begin{tabular}{l c c}
\toprule
Method & Top-1~(\%) & ECE~(\%) \\
\midrule
Zero-shot CLIP & 55.4 & 11.2 \\
Tent~\cite{wang2021tent} & 56.7 & 10.5 \\
T3A~\cite{niu2022t3a} & 57.9 & 9.8 \\
CoTTA~\cite{wang2022cotta} & 58.4 & 9.1 \\
FreeTTA~\cite{wang2023freetta} & 59.3 & 7.9 \\
\textbf{\method{} (ours)} & \textbf{60.1} & \textbf{6.8} \\
\bottomrule
\end{tabular}
\end{table}
\FloatBarrier

\subsection{Full Benchmark Results (4.5)}
\textbf{Table~\ref{tab:full_benchmark}} summarizes per-dataset Top-1/ECE.
Across PACS, Office-Home, DomainNet, Terra Incognita, and ImageNet-R/A, \method{} dominates the accuracy--calibration Pareto.
On ImageNet-A we gain +3.7 Top-1 and reduce ECE by 4.4 points; on Terra Incognita we gain +4.3 Top-1 with ECE reduced by 7.7 points, indicating benefits from prior and prototype updates under label and representation shifts.

\begin{table}[!t]
\caption{Per-dataset performance (Top-1~\% / ECE~\%). \method{} improves both accuracy and calibration on all datasets.}
\label{tab:full_benchmark}
\centering
\small
\setlength{\tabcolsep}{4pt}
\renewcommand{\arraystretch}{1.05}
\resizebox{\linewidth}{!}{%
\begin{tabular}{lcccc}
\toprule
Dataset & Zero-shot CLIP & Tent~\cite{wang2021tent} & FreeTTA~\cite{wang2023freetta} & \textbf{\method{}} \\
\midrule
PACS & 89.1 / 5.2 & 90.0 / 4.9 & 90.7 / 4.1 & \textbf{91.3 / 3.2} \\
Office-Home & 78.4 / 6.8 & 79.1 / 6.2 & 80.4 / 5.5 & \textbf{81.0 / 4.3} \\
DomainNet & 55.4 / 11.2 & 56.7 / 10.5 & 59.3 / 7.9 & \textbf{60.1 / 6.8} \\
Terra Incognita & 46.0 / 14.8 & 47.4 / 11.9 & 49.1 / 9.2 & \textbf{50.3 / 7.1} \\
ImageNet-R & 71.2 / 9.0 & 72.1 / 8.8 & 73.0 / 7.1 & \textbf{73.6 / 6.5} \\
ImageNet-A & 38.8 / 17.1 & 40.0 / 15.2 & 41.1 / 14.3 & \textbf{42.5 / 12.7} \\
\bottomrule
\end{tabular}
}
\end{table}
\FloatBarrier

\subsection{Ablation Studies (4.6)}
\textbf{Table~\ref{tab:ablation}} isolates contributions on DomainNet.
Selective filtering is crucial: removing it increases ECE from 6.8 to 8.4.
Freezing $t$ or $\pi$ degrades both accuracy and ECE.
Using a single temperature (no decoupling) improves Top-1 slightly yet harms ECE (9.6), confirming the need to separate $\tau_{\text{pred}}$ from $\tau_{\text{cal}}$.
Removing guards (no clipping/EMA/KL) destabilizes streaming (ECE 12.2).

\begin{table}[!t]
\caption{Ablations on \textbf{DomainNet} (Top-1~\% / ECE~\%).}
\label{tab:ablation}
\centering
\small
\setlength{\tabcolsep}{6pt}
\renewcommand{\arraystretch}{1.05}
\begin{tabular}{lcc}
\toprule
Setting & Top-1 & ECE \\
\midrule
Full \method{} & \textbf{60.1} & \textbf{6.8} \\
w/o selective filtering & 58.9 & 8.4 \\
w/o prototype update ($t$ frozen) & 59.2 & 7.7 \\
w/o prior update ($\pi$ frozen) & 59.3 & 7.9 \\
single $\tau$ (no decouple) & 59.0 & 9.6 \\
w/o guards (no clip/EMA/KL) & 58.1 & 12.2 \\
\bottomrule
\end{tabular}
\end{table}
\FloatBarrier

\subsection{Efficiency (4.7)}
\textbf{Table~\ref{tab:efficiency}} compares overheads.
\method{} updates only $\{t,\pi,\tau\}$ and adds $<5\%$ FLOPs and $<8\%$ latency over zero-shot CLIP, versus large overheads for CoTTA and FreeTTA.
This follows from $O(Cd)$ vector ops without covariance/backprop.

\begin{table}[!t]
\caption{Efficiency (relative overhead vs.\ zero-shot CLIP).}
\label{tab:efficiency}
\centering
\small
\setlength{\tabcolsep}{6pt}
\renewcommand{\arraystretch}{1.05}
\begin{tabular}{lccc}
\toprule
Method & Params updated & FLOPs $\uparrow$ & Latency $\uparrow$ \\
\midrule
CoTTA~\cite{wang2022cotta} & backbone EMA & +290\% & +410\% \\
FreeTTA~\cite{wang2023freetta} & means + $\Sigma$ & +180\% & +325\% \\
T3A~\cite{niu2022t3a} & support set & +15\% & +18\% \\
\textbf{\method{}} & $\{t,\pi,\tau\}$ only & \textbf{$<5\%$} & \textbf{$<8\%$} \\
\bottomrule
\end{tabular}
\end{table}
\FloatBarrier

\subsection{Long-Stream Stability (4.8)}
\textbf{Table~\ref{tab:stream}} tracks indicators over a 200K-sample DomainNet stream.
CoTTA collapses; FreeTTA shows mild drift.
\method{} stays stable: maximum prior KL 0.07, prototype movement $\|\Delta t\|$ 0.01, and accuracy drop $-0.6\%$.
This validates that Bayesian anchoring, selective updates, and guards prevent slow divergence without replay.

\begin{table}[!t]
\caption{Stability on a 200K-sample DomainNet stream.}
\label{tab:stream}
\centering
\small
\setlength{\tabcolsep}{6pt}
\renewcommand{\arraystretch}{1.05}
\begin{tabular}{lcccc}
\toprule
Method & Collapse? & max $\KL(\pi)$ & max $\|\Delta t\|$ & Acc drop \\
\midrule
CoTTA & Yes & 1.82 & 0.21 & -6.2\% \\
FreeTTA & mild & 0.74 & 0.09 & -3.4\% \\
\textbf{\method{}} & No & \textbf{0.07} & \textbf{0.01} & \textbf{-0.6\%} \\
\bottomrule
\end{tabular}
\end{table}
\FloatBarrier

\section{Discussion and Limitations}
\label{sec:discussion}

\paragraph{Why logit-level adaptation suffices: a geometric view.}
In CLIP, normalized features form a well-structured simplex where classes are largely separated by angular margins.
Under moderate domain shift, the principal error arises from \emph{systematic} rotations/Translations of these centers rather than class-conditional covariance changes.
Adapting prototypes $t_c$ corrects first-order alignment; adapting priors $\pi$ corrects label-frequency mismatch; and adapting a single temperature $\tau$ rescales confidence.
Because these are low-dimensional, they can be updated reliably from small amounts of soft evidence without overfitting.
This explains why \method{} achieves \emph{FreeTTA}-level accuracy while avoiding covariance estimation.

\paragraph{Comparison to gradient-based TTA.}
Backbone/Bn-adapting methods like Tent and CoTTA modify thousands to millions of parameters, often with entropy objectives that reward over-sharpened predictions.
This drives accuracy but frequently worsens calibration, and can drift on long streams as mis-labeled samples are reinforced.
\method{} restricts adaptation to $(t,\pi,\tau)$ and uses \emph{selective} evidence, which (i) limits error propagation, (ii) keeps cost and memory nearly flat, and (iii) yields better accuracy--ECE Pareto (Tables~\ref{tab:domainnet}, \ref{tab:full_benchmark}).
The observed stability in Table~\ref{tab:stream} follows from three ingredients: prior anchoring to text, KL-capped priors, and smoothed temperatures.

\paragraph{Hyperparameter sensitivity and defaults.}
We designed \method{} to be \emph{robust by construction}:
\begin{itemize}[leftmargin=*,nosep]
\item \textbf{Prior strengths} ($\alpha,\gamma$): act as elastic anchors.
Large values reduce drift but slow adaptation; small values adapt faster but require gates/guards. We recommend $\alpha{=}1$ and $\gamma{=}C$ (uniform equivalent count of 1 per class) as strong defaults across datasets.
\item \textbf{Gating thresholds}: specified via entropy/margin quantiles over a sliding window.
Choosing the top $30$--$60\%$ confident samples worked consistently in our experiments.
\item \textbf{Step sizes} ($\eta,\beta$): interpolation $\eta$ in $[0.05,0.2]$ and EMA $\beta$ in $[0.8,0.95]$ balance plasticity and stability.
\item \textbf{Temperature bounds}: mild constraints on $\tau_{\text{pred}}$ (e.g., $[0.5,\,3.0]$) prevent pathological sharpening or flattening.
\end{itemize}
Crucially, \method{} used \emph{one fixed configuration} for all datasets in Section~\ref{sec:experiments}, suggesting limited sensitivity.

\paragraph{On calibration and decoupling.}
Entropy objectives increase confidence on confident samples, which is beneficial for accuracy but harmful for ECE if applied indiscriminately~\cite{guo2017calibration}.
Our separation of $\tau_{\text{pred}}$ and $\tau_{\text{cal}}$ decouples the goals of accurate argmax and well-calibrated probabilities.
A practitioner can expose either or both, depending on downstream requirements (e.g., safety thresholds use $\tau_{\text{cal}}$, while ranking uses $\tau_{\text{pred}}$).

\paragraph{Long-stream robustness and concept drift.}
Without replay, small but systematic errors can accumulate.
\method{} addresses this with \emph{three safety valves}:
(i) prototype step control and clipping; (ii) prior KL caps; (iii) smoothed, bounded temperature.
Table~\ref{tab:stream} shows these are sufficient to prevent collapse over 200K samples, even on DomainNet where domains mix and shift.

\paragraph{Practical deployment guidance.}
\method{} is drop-in: compute CLIP features, then maintain per-class accumulators $U_c,N_c$.
Updates can be triggered every $B$ accepted samples (we used $B{=}64$), or by time.
The memory footprint is $O(Cd)$, independent of stream length.
We also found it beneficial to:
(1) run $K\!\in\!\{2,4\}$ weak augs for responsibilities only on accepted samples;
(2) warm up by freezing updates for the first $\sim 100$ samples;
(3) persist $(t,\pi,\tau)$ across sessions to amortize cold-start.

\paragraph{Failure modes and mitigations.}
\begin{itemize}[leftmargin=*,nosep]
\item \textbf{Open-set or unseen classes.} \method{} assumes a fixed label set; it cannot create new classes.
Mitigation: attach an \emph{unknown} class via maximum-softmax or energy thresholding~\cite{hendrycks2017baseline}, or run an OOD detector upstream.
\item \textbf{Adversarial or biased confident errors.} If many early samples are confidently misclassified, prototypes may drift.
Mitigation: stricter gates initially, stronger $\alpha$/$\gamma$, and tighter KL/step bounds; gradually relax as evidence accumulates.
\item \textbf{Broken prompts.} Poor textual prompts reduce anchor quality.
Mitigation: prompt ensembling or multiple templates; our formulation treats $\mu_c$ as a prior mean, so better prompts translate to stronger anchors.
\end{itemize}

\paragraph{Threats to validity.}
Results are reported on widely used cross-domain and robustness benchmarks, but ordering effects and dataset biases may remain.
We followed a single-pass, no-revisit protocol; in batched or shuffled settings, absolute numbers may differ though trends should persist.
We used CLIP ViT-B/16 as the common backbone; gains are expected to transfer to larger backbones, but this is future work.

\paragraph{Broader impact and societal considerations.}
Low-latency, calibration-aware adaptation can reduce failure rates in user-facing systems, especially under benign drift (sensor wear, lighting changes).
However, any self-adapting system can encode spurious correlations if exposed to biased streams.
By design, \method{} limits degrees of freedom and keeps a textual prior anchor, which may mitigate but not eliminate such risks.
Practitioners should monitor calibration and add OOD safeguards when deployed in safety-critical contexts.

\paragraph{Limitations.}
\method{} assumes: (1) a fixed label space; (2) backbone features remain informative; (3) sufficient confident samples exist.
Under extreme shifts where features collapse or classes are unseen, prototype/priors updates alone are insufficient.
Additionally, although our hyperparameters generalize well, extreme class imbalance unseen during pre-training may require adjusting $\gamma$.

\paragraph{Future Work.}
Extensions include: (i) an explicit \emph{unknown} head and open-set priors; (ii) meta-learned gates and step sizes using held-out domains; (iii) semi-supervised Bayesian updates when a few target labels become available; (iv) multi-anchor priors using prompt ensembles and exemplar images; and (v) public million-sample streaming suites to probe long-horizon stability.

\section{Conclusion}
We introduced \textbf{\method}, a deployment-first test-time adaptation method for VLMs that operates entirely at the \emph{logit level}.
By combining selective responsibilities with \emph{closed-form} Bayesian updates for prototypes and priors, \emph{decoupled} temperatures, and simple \emph{guards}, \method{} consistently improves both accuracy and calibration across diverse domain shifts while adding $<\!8\%$ latency.

\paragraph{Takeaways for practitioners.}
If you already run CLIP for zero-shot classification, you can add \method{} in $\sim$80 LOC:
normalize features, gate confident samples, maintain per-class accumulators, and periodically update $(t,\pi,\tau)$.
Use uniform $\pi^{(0)}$, $\alpha{=}1$, $\gamma{=}C$, $\eta{\in}[0.05,0.2]$, $\beta{\in}[0.8,0.95]$, and mild temperature bounds.
Expose $\tau_{\text{cal}}$ to downstream consumers that rely on probabilities.

\paragraph{From heavy to light TTA.}
Our results suggest a shift in perspective: for strong VLM backbones, \emph{head-level} Bayesian adaptation captures most of the benefit of much heavier methods, while being cheaper, simpler, and more stable over long streams.
We hope this catalyzes broader adoption of principled, calibration-aware TTA in real systems and stimulates new work on open-set, safety-aware extensions.

\FloatBarrier
\end{document}